\documentclass[letterpaper, 10 pt, conference]{ieeeconf}  % Comment this line out if you need a4paper

\IEEEoverridecommandlockouts                              % This command is only needed if 
\usepackage{graphics} % for pdf, bitmapped graphics files
\usepackage{graphicx}
\usepackage[disable]{todonotes}
\usepackage{amsmath} % assumes amsmath package installed
\usepackage{subfigure}
\usepackage{amsfonts}
\usepackage{algorithm,algorithmic}
\usepackage{cite}
\usepackage{mathtools}
\usepackage{multirow}
\usepackage{pifont}
\usepackage{hyperref}

\title{\LARGE \bf
Toward Scalable Visual Servoing Using Deep Reinforcement Learning  and Optimal Control
}

\author{Salar Asayesh$^{1}$, Hossein Sheikhi Darani$^{1}$, Mo Chen$^{1}$, Mehran Mehrandezh$^{2}$ and Kamal Gupta$^{1}$% <-this % stops a space
\thanks{$^{1}$Faculty of Applied Science,
        Simon Fraser University, Burnaby, BC V5A 1S6, Canada, 
        {\tt\small \{sasayesh, hossein\_sheikhi\_darani, mochen, kamal\} at sfu.ca}}%
\thanks{$^{2}$Faculty of Engineering and Applied Science, University of Regina,
        Regina, SK S4S 0A2, Canada,
        {\tt\small mehran.mehrandezh at uregina.ca}}%
}

\renewcommand{\baselinestretch}{.92}
\begin{document}
\maketitle
\thispagestyle{empty}
\pagestyle{empty}

%%%%%%%%%%%%%%%%%%%%%%%%%%%%%%%%%%%%%%%%%%%%%%%%%%%%%%%%%%%%%%%%%%%%%%%%%%%%%%%%
\begin{abstract}

Classical pixel-based Visual Servoing (VS) approaches offer high accuracy but suffer from a limited convergence area due to optimization nonlinearity. Modern deep learning-based VS methods overcome traditional vision issues but lack scalability, requiring training on limited scenes. This paper proposes a hybrid VS strategy utilizing Deep Reinforcement Learning (DRL) and optimal control to enhance both convergence area and scalability. The DRL component of our approach separately handles representation and policy learning to enhance scalability, generalizability, learning efficiency and ease domain adaptation. Moreover, the optimal control part ensures high end-point accuracy. Our method showcases remarkable achievements in terms of high convergence rates and minimal end-positioning errors using a 7-DOF manipulator. Importantly, it exhibits scalability across more than 1000 distinct scenes. Furthermore, we demonstrate its capacity for generalization to previously unseen datasets. Lastly, we illustrate the real-world applicability of our approach, highlighting its adaptability through single-shot domain transfer learning in environments with noise and occlusions. Real-robot experiments can be found at  \url{https://sites.google.com/view/vsls}.
\end{abstract}

%%%%%%%%%%%%%%%%%%%%%%%%%%%%%%%%%%%%%%%%%%%%%%%%%%%%%%%%%%%%%%%%%%%%%%%%%%%%%%%%
\section{INTRODUCTION}

Visual Servoing (VS) is a classical control problem that has been studied for decades and refers to controlling the robot's motion through visual feedback \cite{hutchinson1996tutorial}.
In robotics, VS in fact encompasses a variety of tasks that lie at the intersection of perception and control, from robotic manipulation
to vision-based navigation of mobile robots.%  \cite{6094404, 4084571, 932864}.

Classical Visual Servoing (VS) methods rely on intricate image processing and control steps, requiring substantial feature and model engineering \cite{4015997, hutchinson1996tutorial, 1321161}. They heavily depend on extracting, tracking, and matching 2D or 3D visual features, making them susceptible to performance degradation from input noise, lighting changes, and camera occlusion. Furthermore, these methods assume continuous feature detection, which becomes problematic in practice due to the camera's limited field of view. A newer approach called Direct Visual Servoing (DVS) \cite{collewet2011photometric} avoids the need for feature extraction and matching but has a restricted convergence area compared to classical techniques. However, the optimization is highly non-linear \cite{9689990}.

Current State-Of-The-Art (SOTA) Visual Servoing (VS) systems leverage Deep Neural Networks (DNNs) either in the feedback control loop or in end-to-end processes to enhance feature detection and robustness, albeit at the cost of a slight increase in end effector pose error \cite{AHLIN2016177, kumra2017robotic, 9689990, 8461068, yu2019siamese, 9561488, levine2016end, lee2017learning, Li2019b}. Some strategies employ DNNs in explicit control modules \cite{kumra2017robotic, AHLIN2016177, 9689990}, while others aim to develop control policies through learning methodologies, transforming raw inputs into control commands as an integrated task \cite{8461068, yu2019siamese, 9561488, levine2016end, lee2017learning}. These approaches, however, generally suffer from a lack of scalability, often overfitting to a narrow range of objects during training.

In this study, we introduce a hybrid approach to Visual Servoing (VS) employing both RL-based and DVS methods \cite{collewet2011photometric}. Initially, an RL-based technique guides the end-effector near the DVS convergence area, followed by the activation of DVS for exact positioning. In the RL segment, an unsupervised training of stochastic sequential latent representation takes place, which essentially extracts necessary features from raw input imagery.

This approach facilitates domain adaptation to real-world scenarios through isolated representation learning fine-tuning. It effectively tackles challenges like scalability and partial camera occlusion. Additionally, it exhibits generalization capabilities, delivering consistent performance on unseen scenes. Our experiments, conducted in simulated and real-world environments, validate the efficiency of this novel technique. Notably, it stands as the inaugural scalable modern VS strategy proficient in training across a broad spectrum of objects, marking a pivotal advancement in crafting a scalable and robust solution to prevalent VS challenges.

The contributions of this paper are as follows:
\begin{itemize}
    \item We introduce an innovative hybrid visual servoing (VS) approach employing a goal-conditioned RL-based algorithm with scalable stochastic latent variables, accommodating a wide range of scenes and objects.
    \item We demonstrate that our method attains a high convergence rate and minimal end-positioning error. Moreover, our approach demonstrates consistently high performance even in unseen scenarios.
    \item We show that our method can easily adapt to the real world by single-shot domain transfer on the representation learning part only and validate the effectiveness of our method by implementing it on a 7-DoF manipulator.
    
\end{itemize}

\section{Related Work}
SOTA visual servoing systems employ DNNs either in the feedback control loop of the system \cite{AHLIN2016177, kumra2017robotic,  9689990} or on an end-to-end basis \cite{8461068, yu2019siamese, 9561488, levine2016end, lee2017learning}.
In \cite{AHLIN2016177}, a Convolutional Neural Network (CNN) is trained to detect leaves for a visual servoing task. The CNN was combined with another visual servoing method known as monoscopic depth analysis, which involves comparing two images to determine the location of some feature points in the image relative to the camera in Cartesian space.
\cite{kumra2017robotic} presented a robotic grasp
detection system. The robot is capable of predicting the best grasping pose of a robotic gripper using an RGB-D image of the scene. The method uses two parallel ResNet-50 CNNs, to extract features from RGB and depth in parallel to
produce grasp configurations for the objects in a planar scene. In a recent work \cite{9689990}, a Visual Servoing (VS) framework in latent space was introduced. This framework employs the error in latent space to close the feedback loop for an analytical control module. Their experiments showed that servoing in latent space is an effective approach, offering precise positioning and a more extensive convergence domain compared to other Direct Visual Servoing (DVS) methods. However, it suffers from scalability limitations. 

\cite{8461068} and \cite{yu2019siamese} train a CNN to estimate the relative position error between the current image and the goal image through supervised learning. A major limitation of \cite{8461068} is that the CNN must be retrained for each reference pose. In \cite{yu2019siamese}, current and goal images are fed into a siamese network. The extracted features are then compared at successive layers in order to achieve precise positioning. Siame-SE(3) \cite{9561488}, achieves a high convergence rate in spite of large initial errors but also suffers from scalability. In Siame-SE(3), rather than estimating the positioning error, the camera velocity is directly regressed and learned end-to-end. 

DRL algorithms can in theory employ large-scale deep networks to directly learn policies from pixel inputs \cite{mnih2015human, sac, yarats2021image}. In practice, learning directly from high-dimensional images with a standard end-to-end DRL algorithm can be slow, sensitive to changes in hyperparameters, and data inefficient, since it must address two distinct problems: representation- and policy-learning.
The state-of-the-art attempts to overcome these limitations by leveraging different representation learning methods \cite{srinivas2020curl, lee2020predictive, nair2018visual, lyle2021effect, wayne2018unsupervised, sekar2020planning, slack}.
CURL \cite{srinivas2020curl} proposed a framework to extract high-level features in model-free and model-based RL using contrastive learning and performing off-policy control on top of these features.
Lee et al. \cite{lee2020predictive} showed that capturing the predictive information (mutual information between the past and the future) can be beneficial for RL agents. Their work trains a Soft Actor-Critic \cite{sac} agent from pixels with an auxiliary task that learns a compressed representation of the predictive information of the agent's environment's dynamics using a contrastive loss.

\section{Background}
\label{sec:background}
The innovation of this work stems from the RL-based component of our approach. In the following sections, we will delve into this before discussing the integration of both methods in section. \ref{sec:training_experiments}. Inspired by \cite{slack}, we use representation learning to tackle the VS task, incorporating goal-conditioned multi-task RL, a departure from the approach in \cite{slack}. Our method, resembling \cite{9689990}, necessitates considering past decisions to effectively navigate the latent space in sequential VS tasks.

We outline a goal-conditioned Partially Observable Markov Decision Process (POMDP) defined by the tuple $(\mathcal{X}, \mathcal{Z}, \mathcal{X}^g, \mathcal{Z}^g, \mathcal{A}, \rho, r)$, where elements represent observation, latent variables, goal observation, goal latent variables, action, initial observation distribution, and reward function, respectively. Here, $z \in \mathcal{Z}$ is viewed as the POMDP's unobservable segment, with $p(z_{t+1}|z_t, a_t)$ as the stochastic transition dynamics, assumed to be unknown along with the reward functions and learned through environmental interaction.

For apt representation learning in the VS task, we employed Variational Autoencoders (VAEs) \cite{vae_paper}, an unsupervised approach analyzing input image pixels to extract a condensed and meaningful independent latent space. This technique involves an encoder network $q(z|x)$ transforming pixel space to latent space, and a decoder network $p(x|z)$ reconstructing the input image from its latent representation. Distinct from regular auto-encoders, VAEs leverage stochastic variables grounded on isotropic Gaussian priors for latent space parameters, facilitating control over latent variable distribution and enhancing applicability to large datasets \cite{vae_advantage}. Despite the computational challenges tied to maximizing observation marginal likelihood $p(x)$, VAEs aim to optimize the evidence lower bound (ELBO) for log-likelihood of the observation marginal $\log p(x)$ \cite{vae_paper},
\begin{equation}
    \label{eq:original_elbo}
    %\log p(x) \geq 
    \mathop{\mathbb{E}}_{z \sim q} [\log p(x|z)] - D_{KL}(q(z|x)||p(z)],
\end{equation}
Where $D_{KL}$ denotes the Kullback–Leibler divergence between two distributions, we observe that isolated representation learning fails to furnish the RL policy with essential task information. Hence, we establish $z_t$ as the sequential latent counterpart of observation $x_t$ influenced by transition distribution $p(z_{t+1}|z_t, a_t)$, enriching the information derived from previous observation and action \cite{slack}. Accordingly, we target to maximize the distribution $\log p(x_{1:\tau+1}|a_{1:t})$, akin to \eqref{eq:original_elbo}, as defined by:
\begin{equation}
\label{eq:elbo_loss_sequential}
\begin{split}
    %\log p(x_{1:\tau}|a_{1:\tau}) \geq 
    \mathop{\mathbb{E}}_{z_{1:\tau+1} \sim q}[\sum_{t=0}^\tau \log p(x_{t+1}|z_{t+1} \\
    - D_{KL}(q(z_{t+1}|x_{t+1}, z_t, a_t)\; ||\; p(z_{t+1}|z_t, a_t)]
\end{split}
\end{equation}
where $p(x_t|z_t)$ is the decoder model, $p(z_{t+1}|z_t, a_t)$ is the prior model, $p(z_1)$ is the initial prior, $q(z_{t+1}|x_{t+1}, z_t, a_t)$ is the variational posterior.

Following the work on variational inference \cite{variational_inference}, and SLAC \cite{slack}, we incorporate the control into the inference problem and define the problem as maximizing the following goal-conditioned marginal likelihood:
\begin{equation}
    \label{eq:goal_cond_marginal}
    p(x_{1:\tau+1}, \mathcal{O}_{\tau+1:T}|a_{1:\tau}, x^g)
\end{equation}
where $\mathcal{O}$ is a binary optimality variable, with
\begin{equation}
\label{eq:optimality_def}
    p(\mathcal{O}_t=1) = \exp (r_t).
\end{equation}

Eq. \eqref{eq:goal_cond_marginal} represents both representation learning and policy learning by maximizing the likelihood of observed data and maximizing the policy behaviour over future steps, respectively in a single objective. Similar to \cite{slack}, we factorized the latent variable $z_t$ into two stochastic latent variables $(z_t^1, z_t^2)$ to make the learning process more expressive. Similar to \eqref{eq:elbo_loss_sequential}, we maximize ELBO over \eqref{eq:goal_cond_marginal} instead of directly maximizing the \eqref{eq:goal_cond_marginal},
\begin{equation}
    \label{eq:elbo_final}
    \begin{split}
        \mathop{\mathbb{E}}_{(z_{1:T}^1, z_{1:T}^2, a_{\tau+1:T}, z^g)\sim q}
        \Bigg [\sum_{t=0}^T(\log p(x_{t+1}|z_{t+1}^1, z_{t+1}^2)\\
        - D_{KL}(\log q(z_{t+1}^1|x_{t+1}, z_t^2, a_t)||\log p(z_{t+1}^2|z_{t+1}^1, a_t))\\
        +\sum_{t=\tau+1}^t \Big( r_t - \log p(a_t) 
        - \log\pi (a_t|x_{1:t}^1, a_{1:t-1}, x^g \Big) \Bigg],
    \end{split}
\end{equation}
The first summation term in \eqref{eq:elbo_final} is similar to \eqref{eq:elbo_loss_sequential} trying to maximize the likelihood of $x_{1:\tau+1}$ in \eqref{eq:goal_cond_marginal}, however, with two sequential latent variable $(z_1, z_2)$. The second summation term corresponds to maximizing the likelihood of the optimality variable in \eqref{eq:goal_cond_marginal}. For more details please refer to \cite{slack} and \cite{variational_inference}. 

Eq. \eqref{eq:elbo_final} has an excellent interpretation that the first part corresponds to representation learning and the second part corresponds to VS task objective part which is similar to the maximum entropy RL objective \cite{sac}.

\section{Method}
\subsection{Observation and Action space}
\label{subsec:observation_action_subsec}
The observation is designated as a history of image-action pairs $x_t = (o_t, o_{t-1}, ... o_{t-T_L}, a_{t-1}, a_{t-2}, ... a_{t-T_L})$, where $o_t$ is the camera-captured image at time $t$. This data, detailed further in \ref{subsec:rl_algorithm}, is condensed into feature and latent components for control policy learning.

The policy network predicts short-term navigational goals using the camera's Cartesian space displacement as the action space, defined stochastically per \eqref{eq:action_space}. This action space choice enhances real-robot implementation, marrying policy network predictions with low-level controller efficiency for precise goal navigation.
\begin{equation}
    \label{eq:action_space}
    a_t = (\delta_x, \delta_y, \delta_z, \delta_\text{roll}, \delta_\text{pitch}, \delta_\text{yaw})
\end{equation}

Each $\delta_w$ follows a specific normal distribution, guiding the low-level controller to attain goals within three iterations. The policy network's predictions maintain bounds to ensure goal attainability within three iterations by the low-level controller, assuming feasibility.

\subsection{Reward Function}
The VS task can alternatively be viewed as reaching a specific 6D pose in the manipulator's workspace to capture the target image. We presume knowledge of the current and target end-effector/camera poses during training. The camera pose is defined as follows:
\begin{equation}
    \label{eq:camera_state}
    p_t = (p_t^T, p_t^R)
\end{equation}
where $p_t^T$ denotes the camera translation in Cartesian coordinate and $p_t^R$ denotes the camera rotation in quaternion format at time $t$. Furthermore, the reward function is formulated as:
\begin{equation}
    \label{reward_function}
    r(x_t, a_t) =
    \begin{cases}
    r_g  \;\;\;\;\;\;\;\;\;\;\;\;\;\;\;\;\;\;\;\;\;\;\ \text{ if }d_\text{goal}^t < $c$\\
    d_\text{goal}^{t-1} - d_\text{goal}^t   \;\;\;\;\;\;\;\;\;\text{ otherwise }
    \end{cases}
\end{equation}
where $c$ denotes a threshold value for reaching the goal and $d_\text{goal}^t$ is a defined distance function from the current to goal configuration given by:
\begin{equation}
\label{eq:distance_fcn_reward}
        d_\text{goal}^t = c_1 ||p_t^T - p_{\text{goal}}^T|| + c_2|(p_\text{goal}^R)^\top  [p_t^{R}]^{-1}|
\end{equation}
Where $c_1, c_2$ are constant scaling factors, \eqref{eq:distance_fcn_reward} presents two terms: the first measures the Euclidean distance between the current and target positions in Cartesian space, and the second gauges the rotational magnitude from the present to the target pose, with both scaled to a [0, 1] range considering the manipulator's workspace boundaries.

\subsection{Curriculum Learning}
\label{subsec:curriculum}
To enhance the learning process, we reference \cite{9689990} for our curriculum algorithm and apply a reverse expansion on goal distribution. We assume a planar object, differing from \cite{9689990, 9561488}, with bounded perturbations in all 6 DOFs.

Training covers two scenarios: look-at and screw motion. For look-at, goal positions $P$ are taken from a box $[d_1, d_2, d_3]$ within the manipulator's workspace. Camera angles focus on a point within the object's center radius $r$, with a random roll angle from $\mathcal{N}(0,\sigma)$. For screw motion, the end-effector moves and rotates only along the x-axis. Movements and rotations come from ($d1, \theta_{max}$), ensuring goal $x$ and initial images align well. After verifying the goal's reachability using the IK solver, we assess the policy on that goal. Goals are only added if the policy fails the task. We observed that not training on achievable goals enhances convergence. We have $N$ curriculum stages, each with $K$ goals. Post $K$ goals, curriculum variables like box dimensions, scene perturbations, and look-at parameters increase. To avoid forgetting, solved goals are randomly revisited.

\subsection{Proposed Architecture and Training Algorithm}
\label{subsec:rl_algorithm}
\begin{figure}
    \centering
    \includegraphics[scale=0.31]{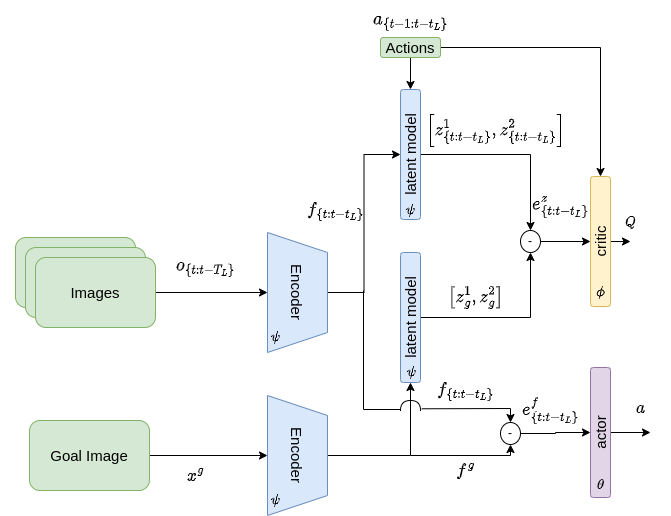}
    \caption{The proposed architecture}
    \label{fig:critic_architecture}
    \vspace{-6mm}
\end{figure}

Fig. \ref{fig:critic_architecture} shows the proposed architecture. 
The system architecture is divided into representation and manipulation units. The latter contains an actor-network (purple box) with parameters $\theta$ and a critic network (yellow box) defined by $\phi$. The representation unit houses an encoder network, a latent model, and a decoder network (depicted in blue and parameterized by $\psi$). While the decoder network is not shown in Fig. \ref{fig:critic_architecture} for simplicity, it is an integral part of the system.
Inputs, comprising a history of image observations ($o_{t:t-t_L}$) and the goal image ($x_g$), are initially processed by the encoder network to generate feature vectors for each image ($f_{t:t-t_L}$) and the goal ($f^g$). These vectors, paired with a history of actions, help estimate the latent variables ($z_{t:t-t_L}^1, z_{t:t-t_L}^2, z_g^1, z_g^2$). Both the encoder and latent model networks are utilized for image history and goal image processing but are shown separately in Fig. \ref{fig:critic_architecture} for clarity.
It should be noted that goal latent variables ($z_g^1, z_g^2$) are non-sequential and derived from the primary posterior estimation of the goal observation, given that the goal observation is selected from the goal distribution. 

We compute the discrepancy in image history observations between the target and provided images, both in the latent ($e_{1:t}^z$) and feature spaces($e_{1:t}^f$). An asymmetric actor-critic \cite{pinto2017asymmetric} is utilized, feeding the critic with latent space error and the actor with feature space error. The distinction is that feature vectors aren't bound by the Markov property. This design choice enhances robustness in real-robot domain transitions and streamlines algorithm execution.
\begin{figure}
    \centering
    % \includegraphics{}
    % \caption{Caption}
    \label{fig:empty}
    \vspace{-2mm}
\end{figure}
We train the critic network (parameterized by $\phi$) using the following loss:
\begin{equation}
    \label{eq:critic_loss}
    J_\phi = \mathop{\mathbb{E}}_{(e_{1:t+1})\sim (q^1_\psi, q^2_\psi)} \Bigg[\frac{1}{2}\Big(Q_\phi(e_t)- (r_t + \gamma V_{\Bar{\phi}}(e_t, o_t, x^g) \Big )^2 \Bigg]
\end{equation}
where $\Bar{\phi}$ is the delayed target network that is given by
\begin{equation}  
    \label{eq:target_value_function}
    \begin{split}
        V_{\phi}(e_t, o_t, x^g) =  \mathop{\mathbb{E}}_{a_{t+1} \sim \pi_\theta} \Big [Q_\phi(e_t)- \alpha \log \pi_\theta(a_{t+1}|o_t, x^g) \Big].
    \end{split}
\end{equation}
We train the policy (actor) network (parameterized by $\theta$) by maximizing the following objective \cite{sac}:
\begin{equation}
    \label{eq:policy_loss}
    \begin{split}
        J_\theta = \mathop{\mathbb{E}}_{e_{1:t+1} \sim (q^1_\psi, q^2_\psi)}
        \Bigg [\mathop{\mathbb{E}}_{a_{t+1} \sim \pi_\theta} \Big[\alpha \log \pi_\theta(a_{t+1}|o_{t+1}, x^g) \\
        - Q_\phi(e_{t+1}) \Big] \Bigg]
    \end{split}
\end{equation}
Note that in \eqref{eq:policy_loss}, the $J_\theta$ does not depend on the representation model parameters $\psi$. Based on \eqref{eq:elbo_final}, the latent space parameters ($\psi$) are optimized to minimize the ELBO loss using the reparameterization trick and
\begin{equation}
    \label{eq:elbo_loss}
    \begin{split}
        J_\psi =  \mathop{\mathbb{E}}_{(z^1_{1:t+1}, z^2_{1:t+1})\sim (q^1_\psi, q^2_\psi)} \Bigg [\sum_{t=0}^{\tau} - \log p_\psi(x_{t+1}|z_{t+1}^1, z_{t+1}^2)\\
        +D_{KL}(\log q^1_\psi(z_{t+1}^1|x_{t+1}, z_t^2, a_t)||\log q^2_\psi(z_{t+1}^2|z_{t+1}^1, a_t)) \Bigg].
    \end{split}
\end{equation}

Algorithm \ref{alg:training} outlines the training procedure. We initialize network parameters and choose a random first-stage goal. After $M$ training steps, we evaluate the policy; if it exceeds $R_t$, the goal is achieved. This repeats until $K$ goals are met in stage $N$. Meeting the $K$ goals advances the stage and restarts the process.
 \subsection{Data Augmentation}
To enhance domain transfer robustness for VS challenges, we employ comprehensive data augmentation strategies using a plethora of realistic images. Initially, we utilize a pre-trained ResNet-18 as the encoder, cited in \cite{resnet}. Following this, we introduce CutOut data augmentation to simulate partial occlusion by blackening random rectangular portions of images. Lastly, we impose various alterations such as noise addition, lighting adjustments, color bit reduction, and grayscale conversions to fortify the dataset.
 \subsection{Single-Shot Domain Transfer}

In section. \ref{sec:results_discussion}, we discuss performance dips when transitioning from simulation to real-robot application, despite our diligent data augmentation and simulation design. This occurs due to unaccountable variations between the simulated and real-robot domains. Our proposed solution is the single-shot domain transfer algorithm, which fine-tunes the representation learning part. In this method, we first run the algorithm for one episode and record the observed data on the real robot. Then we perform a few fine-tuning steps on the representation learning part, according to Line 11 in Algo. \ref{alg:training}. This allows adjustments without altering the manipulation learning component, facilitating real-world applications where detailed states and reward functions may be inaccessible.
  
 \begin{algorithm}
 \caption{VSLS algorithm training loop}
 \label{alg:training}
 \begin{algorithmic}[1]
 \REQUIRE 
  initialize parameters \(\phi, \theta, \psi, lr_\phi, lr_\theta, lr_\psi\)\\
  initialize buffer $D$ and goal $x^g$ \\
  \WHILE{Not converge}
  \STATE{$R \leftarrow \text{EvaluatePolicy}(\pi_\theta, \text{goal})$}
  \IF{$R \ge R_t$}
  \STATE{$x^g \leftarrow \text{Curriculum}()$}
  \ENDIF
  \FOR{step $s$ in \{1, ... , M\}}
  \STATE{$a_t \sim \pi_\theta(a_t|o_{1:t}, x^g)$}
  \STATE{$r_t, o_{t+1} \leftarrow \text{rollout action}$}
  \STATE{$D\leftarrow D \cup (a_t, r_t, o_{t+1}, x^g)$}
  \STATE{$\text{sample mini batch } b^i \sim D$}
  \STATE{$\psi \leftarrow \psi - lr_\psi \nabla_\psi J_\psi(b^i) \textit{ Optimize } \eqref{eq:critic_loss}$}
  \STATE{$\theta \leftarrow \theta - lr_\theta \nabla_\theta J_\theta(b^i) \textit{ Optimize }\eqref{eq:policy_loss}$}
  \STATE{$\phi \leftarrow \phi - lr_\phi \nabla_\phi J_\phi(b^i) \textit{ Optimize }\eqref{eq:elbo_loss}$}
  \ENDFOR
  \ENDWHILE
 \end{algorithmic}
 \end{algorithm}
\section{Training and Experiments}
\label{sec:training_experiments}
We implemented the RL component of our algorithms using PyTorch and carried out all training procedures on the Isaac Gym simulator \cite{makoviychuk2021isaac}. Isaac Gym is a high-performance, GPU-accelerated learning platform designed for training policies. This environment facilitates direct communication between physics simulations and neural network policy training through the seamless transfer of data from physics buffers to PyTorch tensors on GPUs.
To accelerate the domain adaptation process for real robots and acquaint the algorithm with a wide array of photo-realistic images, we employed datasets from ImageNet \cite{imagenet}. These images were integrated as textures of objects within the Isaac Gym simulation environment.
In terms of architecture, the encoder leverages a pre-trained ResNet-18 model \cite{resnet}, while the decoder employs a custom-designed structure. 

As outlined in section. \ref{sec:background}, we conceptualized our problem within a finite-time horizon, adopting a finite-episodic method for task resolution. The task is deemed complete when the average translation error dips below 3 cm and the average orientation error is less than $2^{\circ}$. These thresholds were chosen because they allow the end-effector to enter the convergence area of Direct Visual Servoing (DVS) \cite{collewet2011photometric, 9561488}. We apply DVS to achieve precise end-point positioning subsequently using the visp \cite{visp} code base.
Training the algorithm necessitated 4-5 days of computational time, spanning two million iterations on a system equipped with an NVIDIA RTX-3090 GPU, an Intel\textsuperscript{\textregistered} Core\textsuperscript{\tiny{TM}} i7-9700K processor, and 64 GB of RAM.
For the real robot experiments, we utilized a Kinova Gen-3 robot equipped with 7 degrees of freedom, complemented by a RealSense RGBD camera mounted in an eye-in-hand configuration\footnote{Note that the depth data was not utilized in this study.}.
\begin{figure*}
    \centering
    % \includegraphics{}
    % \caption{Caption}
    \label{fig:empty}
    \vspace{-2mm}
\end{figure*}

\begin{table*}[]
\centering
\begin{tabular}{|l|l|ll|ll|}
\hline
\multirow{2}{*}{} & \multirow{2}{*}{Scalibility} & \multicolumn{2}{c|}{look-at}                                                                                                                                  & \multicolumn{2}{c|}{screw motion}                                                                                                                             \\ \cline{3-6} 
                  &                              & \multicolumn{1}{c|}{\begin{tabular}[c]{@{}c@{}}success rate\\ \%\end{tabular}} & \multicolumn{1}{c|}{\begin{tabular}[c]{@{}c@{}}End error\\ cm,$^{\circ}$\end{tabular}} & \multicolumn{1}{c|}{\begin{tabular}[c]{@{}c@{}}success rate\\ \%\end{tabular}} & \multicolumn{1}{c|}{\begin{tabular}[c]{@{}c@{}}End error\\ cm,$^{\circ}$\end{tabular}} \\ \hline
DVS               & NA                           & \multicolumn{1}{l|}{72}                                                        & 0.2, 0.8                                                                     & \multicolumn{1}{l|}{64}                                                        & 0.1,0.5                                                                      \\ \hline
S-Se(3)           & 1 scene                      & \multicolumn{1}{l|}{100}                                                       & 0.92, 0.67                                                                   & \multicolumn{1}{l|}{N/A}                                                        & N/A                                                                     \\ \hline
AE VS             & 1 scene                      & \multicolumn{1}{l|}{94.6}                                                      & 0.003, 0.003                                                                 & \multicolumn{1}{l|}{91}                                                        & 0.006, 0.006                                                                 \\ \hline
Ours              & 1004 scenes                  & \multicolumn{1}{l|}{96}                                                        & 0.2, 0.56                                                                    & \multicolumn{1}{l|}{94}                                                        & 0.24, 0.3                                                                    \\ \hline
Ours (RL only)    & 1004 scenes                  & \multicolumn{1}{l|}{97.2}                                                      & 2, 1.7                                                                       & \multicolumn{1}{l|}{93.2}                                                      & 1.4, 1.8                                                                     \\ \hline
\end{tabular}
\caption{comparison with other direct visual servoing methods}
\label{table1:comparison}
\vspace{-10mm}
\end{table*}

\section{Results and Discussions}
\label{sec:results_discussion}
In this section, we explore the results, discussing the following points:
\begin{itemize}
    \item Performance analysis and comparison to other SOTA methods
    \item  Evaluating the generalizability fostered through representation learning in sequential stochastic latent models
    \item Transferring the policy to the real domain
\end{itemize}

\textbf{Performance}
To compare our method's convergence rate, we first train our method with a dataset of 1004 different scenes and objects from ImageNet\cite{image_net} by sampling one image from each class of ImageNet and supplementing this with a custom image dataset.
Adapting the test scenarios proposed in \cite{9689990}, we considered two cases: Look-at and Screw-Motion. In the first test case, the end-effector carrying the camera randomly moves to a point sampled from the robot's workspace and attempts to look at a random point within in the $r$-radius neighbourhood of the image center. The roll angle of the end-effector is randomly selected. Due to physical robot and experiment limitations, we sampled pose from a box [$0.75$ m, $0.75$ m, $0.6$ m] expanded from the home position of the robot equally in all directions. We calculate the pitch and yaw angles such that the camera look-at the $10$ cm neighbourhood of the center of the image, and the roll angle is kept small. In the latter, we assume the end-effector has a displacement in the range of [$-0.25$ m, $0.25$ m] in the x-axis and rotation in the range of 70$^{\circ}$. For each scenario, we randomly generate 500 samples and ensure that the sampled points are feasible for the manipulator by simply replacing any that are not feasible. The starting average errors are $21.12 \text{ cm} \pm 4 \text{ cm}$ and $17.7^{\circ} \pm 4^{\circ}$. 

We evaluate and compare our method with other direct VS approaches, namely DVS \cite{caron2013photometric} as the baseline classical approach, AEVS \cite{9689990} and Siamese-SE(3) \cite{9561488} as SOTA methods. The DVS is implemented in our physics-based simulator using ViSP \cite{visp} and ROS.

Since the AEVS and Siamese-SE(3) implementations are not open-source, we endeavoured to replicate the experiments as closely as possible, utilizing their reported results for comparison \cite{9689990}. It should be noted that comparison with \cite{9561488, {9689990}} is qualitative and may not be precisely accurate. Our objective is to demonstrate that our method can achieve a relatively acceptable end-positioning error and convergence success rate, even with a scalability magnitude of 1000, compared to other methods.
Therefore, there could be discrepancies in the average initial errors between these methods and ours. For instance, the initial average error for Siamese-SE(3) \cite{9561488} is reported to be $12.5 \text{ cm} \pm 7.5 \text{ cm}$ and $17.5^{\circ} \pm 12^{\circ}$. 
Furthermore, the results for AEVS and Siamese-SE(3) were not generated based on a robot-affixed camera operating within a physics-based simulation; rather, they were produced solely for training within a single-scene context.

\begin{figure}%
    \centering
    \subfigure[]{{\includegraphics[width=2.6cm]{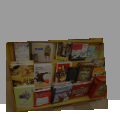} }}%
    % \qquad
    \subfigure[]{{\includegraphics[width=2.6cm]{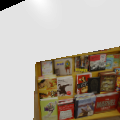} }}%
    \subfigure[]{{\includegraphics[width=2.6cm]{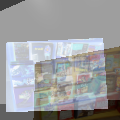} }}%
    \newline
    \subfigure[]{{\includegraphics[width=1.6cm]{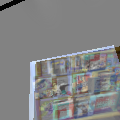} }}%
    \subfigure[]{{\includegraphics[width=1.6cm]{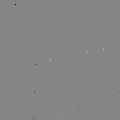} }}%
    \subfigure[]{{\includegraphics[width=3.6cm]{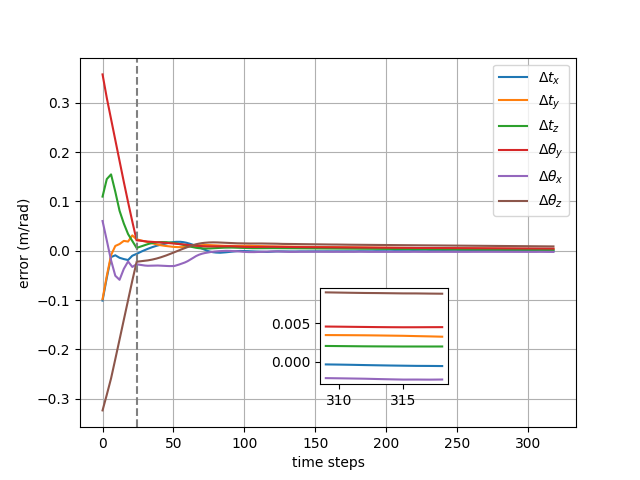} }}%
    \caption{One of the generated random trajectories (a) initial Image; (b) target Image; (c) initial Image difference; (d) Image difference at the end of RL running; (e) final image difference; (f) pose difference}%
    \label{fig:generated_traj}%
    \vspace{-5mm}
\end{figure}

Fig. \ref{fig:generated_traj} displays one of the generated trajectories. Initially, the RL algorithm is activated, guiding the end-effector within close proximity to the desired position. Subsequently, DVS is deployed to enhance end-point accuracy. These two phases are delineated by a vertical grid line in Fig. \ref{fig:generated_traj}-f.

Table \ref{table1:comparison} compares our method with other direct VS methods. For scalability assessment, we took into account the number of different images utilized during both the training and test phases. Since the DVS \cite{collewet2011photometric} is not learning-based, it is scalable and not limited to the size of the dataset.
The data indicates that our approach can significantly expand the convergence area of DVS while achieving a relatively low end-positioning error. Additionally, we report solely on the end-positioning error derived from the RL portion of our algorithm. It is important to note that both the DVS and our full method exhibit higher end-positioning errors compared to the results reported in \cite{9561488, 9689990}. This discrepancy might be attributed to our utilization of a physics-based simulator and a camera mounted on a 7-DOF manipulator, in contrast to the cited works, which did not leverage physics-based simulation and allowed for unrestricted camera movement within the environment. 

\begin{figure}%
    \centering
    \subfigure[]{{\includegraphics[width=2.6cm]{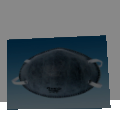} }}%
    % \qquad
    \subfigure[]{{\includegraphics[width=2.6cm]{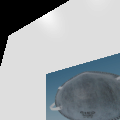} }}%
    \subfigure[]{{\includegraphics[width=2.6cm]{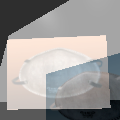} }}%
    \newline
    \subfigure[]{{\includegraphics[width=2.6cm]{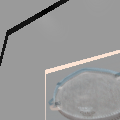} }}%
    \subfigure[]{{\includegraphics[width=3.9cm]{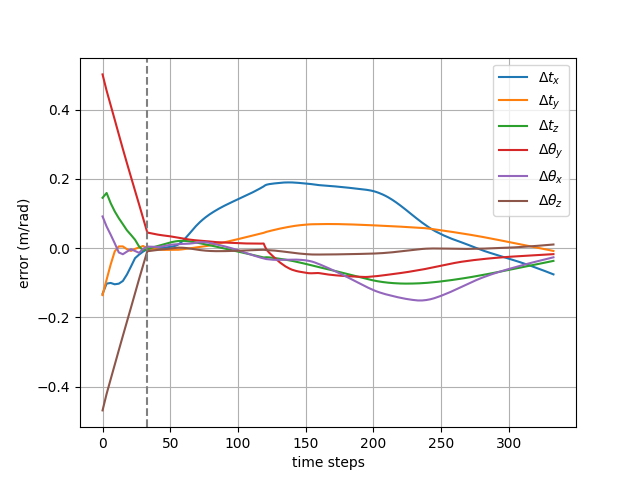} }}%
    \caption{One of the failure cases after applying DVS (a) initial Image; (b) target Image; (c) initial Image difference; (d) Image difference at the end of RL running; (e) pose difference}%
    \label{fig:failure_cases}%
    \vspace{-5mm}
\end{figure}

Interestingly, our complete method experiences a minor decrease in success rate compared to utilizing only the RL component during the AE test. This decrease can be traced back to the complexity of the task and the inadequate presence of distinguishable features in the scenes - a situation where the DVS falls short, even when the RL segment manages to meet an acceptable threshold before handing off to the DVS. This failure is depicted in Fig. \ref{fig:failure_cases}, with Fig. \ref{fig:failure_cases}.e highlighting the exact moment the DVS is activated, indicated by the vertical grid line. Contrarily, when dealing with screw motions, the use of DVS serves to improve the convergence success rate of the entire methodology.

\textbf{Generalizability}
To evaluate the generalizability of our method amid the recognized low generalizability in end-to-end RL, we deployed the trained policy and tested it against a new dataset. For the new data set, we chose one of the ImageNet classes -dog class:  a class that was only minimally represented in the training dataset by only one image- and sourced 1000 unseen images from Microsoft COCO \cite{microsoft-coco} in the same class. The outcomes are summarized in Table \ref{table2:unseen_data}. Given that the DVS component is not learning-based and remains agnostic to the dataset, we have limited the report to the results derived from the RL segment for a concise comparison. Despite historical challenges in end-to-end RL generalizability, our approach exhibited minimal performance dip, suggesting an effective feature extraction without data memorization, thanks to the implementation of separate representation learning in our sequential stochastic latent model. This is further corroborated by the maintenance of a relatively consistent end-positioning error margin when compared to the observed dataset.

\begin{table}[H]
\centering
\begin{tabular}{|l|cl|cl|}
\hline
\multirow{2}{*}{} & \multicolumn{2}{c|}{look-at}                                                                                                                                  & \multicolumn{2}{c|}{screw motion}                                                                                                                             \\ \cline{2-5} 
                  & \multicolumn{1}{c|}{\begin{tabular}[c]{@{}c@{}}success rate\\ \%\end{tabular}} & \multicolumn{1}{c|}{\begin{tabular}[c]{@{}c@{}}End error\\ cm, $^{\circ}$ \end{tabular}} & \multicolumn{1}{c|}{\begin{tabular}[c]{@{}c@{}}success rate\\ \%\end{tabular}} & \multicolumn{1}{c|}{\begin{tabular}[c]{@{}c@{}}End error\\ cm, $^{\circ}$\end{tabular}} \\ \hline
Ours(RL only)    & \multicolumn{1}{l|}{93 \%}                                                     & 2, 1.7                                                                       & \multicolumn{1}{l|}{89 \%}                                                     & 1.4, 2.1                                                                     \\ \hline
\end{tabular}
\caption{Performance of proposed approach on unseen dataset}
\label{table2:unseen_data}
\vspace{-5mm}
\end{table}

\begin{figure}%
    \centering
    \subfigure[]{{\includegraphics[width=2.6cm]{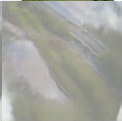} }}%
    \subfigure[]{{\includegraphics[width=2.6cm]{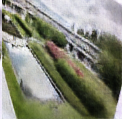} }}%
    \caption{Reconstructed Image from Latent Space; (a) Before Single-Shot transfer; (b) After Single-Shot transfer}%
    \label{fig:reconstructed_sequence}%
    \vspace{-5mm}
\end{figure}

\begin{figure}%
    \centering
    \subfigure[]{{\includegraphics[width=2.6cm]{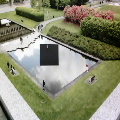} }}%
    % \qquad
    \subfigure[]{{\includegraphics[width=2.6cm]{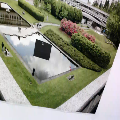} }}%
    \subfigure[]{{\includegraphics[width=2.6cm]{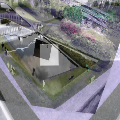} }}%
    \newline
    \subfigure[]{{\includegraphics[width=2.6cm]{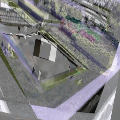} }}%
    \subfigure[]{{\includegraphics[width=2.6cm]{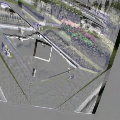} }}%
        \newline
    \subfigure[]{{\includegraphics[width=4cm]{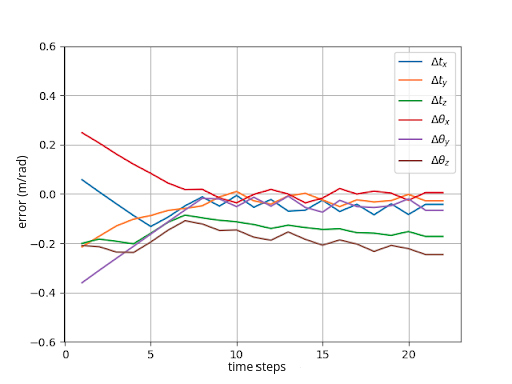} }}%
    \subfigure[]{{\includegraphics[width=4cm]{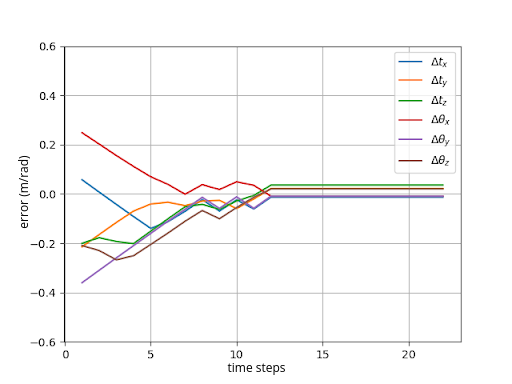} }}%
    \caption{Real robot experiment with occlusion; (a) initial Image; (b) target Image; (c) initial Image difference; (d) final Image difference before single-shot transfer; (e) final image difference after single-shot transfer; (f) pose difference before single-shot transfer (g) pose difference after single shot transfer}%
    \label{fig:real-experiment}%
    \vspace{-5mm}
\end{figure}

\textbf{Real Robot Experiment:}
We finally deployed our method to a real 7 DOF Kinova Gen3 robot to test in real scenes, leveraging a scene drawn from our training dataset \footnote{Images taken from Simon Fraser University library with permission}. Note that we restrict our demonstration to the results derived from the RL segment of our algorithm. This is because the RL component is learning-based and thus is susceptible to sim2real issues. To underscore the pivotal role of single-shot domain transfer, we utilized the decoder network $p(x_t|z^1_t, z^2_t)$, as defined in Eq. \eqref{eq:elbo_final}, to reconstruct real experiment images from the encoded latent space. This process, visually represented in Fig. \ref{fig:reconstructed_sequence}, significantly reduces reconstruction error and requires under two minutes for fine-tuning on an NVIDIA RTX-2080 GPU. Fig. \ref{fig:real-experiment} details an entire experiment, demonstrating the proficiency of our method in real-world scenarios and highlighting the enhanced performance achieved through the single-shot domain transfer, even when faced with partial occlusion challenges. The figure clearly illustrates that the application of single-shot transfer enables successful task convergence, confirming its effectiveness in a real-world setting.

\section{CONCLUSIONS}

This paper introduces a hybrid VS solution grounded on RL-based goal-conditioned visual servoing algorithm which leverages sequential latent space representation. Our method, which harmoniously blends traditional VS strategies with modern machine learning techniques, emerges as a pivotal advancement towards addressing the longstanding issues in visual servoing with a higher degree of robustness and scalability.


%%%%%%%%%%%%%%%%%%%%%%%%%%%%%%%%%%%%%%%%%%%%%%%%%%%%%%%%%%%%%%%%%%%%%%%%%%%%%%%%

%%%%%%%%%%%%%%%%%%%%%%%%%%%%%%%%%%%%%%%%%%%%%%%%%%%%%%%%%%%%%%%%%%%%%%%%%%%%%%%%

%%%%%%%%%%%%%%%%%%%%%%%%%%%%%%%%%%%%%%%%%%%%%%%%%%%%%%%%%%%%%%%%%%%%%%%%%%%%%%%%

\clearpage

\bibliographystyle{IEEEtran}
\bibliography{main}
\end{document}